\documentclass[journal]{IEEEtran}
\usepackage[colorlinks=false]{hyperref}

\usepackage{tabularx}
\usepackage{graphicx}

\usepackage{cite}

\usepackage{accents}
\usepackage{textcomp}
\usepackage{multirow}
\usepackage{booktabs}
\usepackage{float}

\usepackage{amsfonts}

\ifCLASSINFOpdf
\else
\fi

\usepackage[numbers]{natbib}

\usepackage[colorinlistoftodos]{todonotes} 

\definecolor{orange}{HTML}{FFC17D}
\definecolor{blue}{HTML}{7BABFF}
\definecolor{green}{HTML}{A1D68B}
\definecolor{lightgray}{HTML}{E8E8E8}

\begin{document}

\title{Explaining Credit Risk Scoring through Feature Contribution Alignment with Expert Risk Analysts}

%
%

\author{Ayoub~El-Qadi,
        Maria~Trocan, 
        Thomas~Frossard and
        Natalia~Díaz-Rodríguez 
\thanks{A. El Qadi is with Sorbonne University, ENSTA, Institut Polytechnique Paris, and Tinubu Square, 169 Quai de Stalingrad, 92130 Issy-les-Moulineaux, France.
\href{mailto:aei.ext@tinubu.com}{aei.ext@tinubu.com}
\href{mailto:ayoub.el_qadi_el_haouari@etu.sorbonne-universite.fr}{ayoub.el\_qadi\_el\_haouari@etu.sorbonne-universite.fr}.}
\thanks{M. Trocan is with LISITE, ISEP, Institut Supérieur d'Électronique de Paris,  10, rue de Vanves, 92130 Issy-les-Moulineaux, France.
\href{mailto:maria.trocan@isep.fr}{maria.trocan@isep.fr}.}
\thanks{T. Frossard is with Tinubu Square, 169 Quai de Stalingrad, 92130 Issy-les-Moulineaux, France.
\href{mailto:tfd@tinubu.com}{tfd@tinubu.com}.}
\thanks{N. Díaz-Rodríguez is with U2IS, ENSTA Paris, Institut Polytechnique Paris, Inria Flowers Team, 828 boulevard des Maréchaux, 91762 Palaiseau, France. \href{mailto:natalia.diaz@ensta-paris.fr}{natalia.diaz@ensta-paris.fr}}}
\maketitle
\begingroup\renewcommand\thefootnote{\textsection}

\begin{abstract}
Credit assessments activities are essential for financial institutions and allow the global economy to grow. Building robust, solid and accurate models that estimate the probability of a default of a company  is mandatory for credit insurance companies, specially when it comes to bridging the trade finance gap. Automating the risk assessment process will allow credit risk experts to reduce their workload and focus on the critical and complex cases, as well as to improve the loan approval process by reducing the time to process the application. The recent developments in Artificial Intelligence are offering new powerful opportunities. However, most AI techniques are labelled as black-box models due to their lack of explainability. For both users and regulators, in order to deploy such technologies at scale, being able to understand the model logic is a must to grant accurate and ethical decision making. In this study, we focus on companies credit scoring and we benchmark different machine learning models. The aim is to build a model to predict whether a company will experience financial problems in a given time horizon.  We address the black box problem using eXplainable Artificial Techniques --in particular, post-hoc explanations using SHapley Additive exPlanations. We bring light by providing an expert-aligned feature relevance score highlighting the disagreement between a credit risk expert and a model feature attribution explanation in order to better quantify the convergence towards a better human-aligned decision making.
\end{abstract}

%
\IEEEpeerreviewmaketitle

\section{Introduction}
%
%
%
%
\IEEEPARstart{F}{or} banks and more generally financial and insurance institutions credit risk assessments activities are the engine of their industry. The approach to evaluate the risk depends on the type of counterpart (i.e., the part that receives the credit) and how the risk has been assessed. Governments  or corporate (i.e., companies publicly traded), small and medium companies (i.e., companies publicly traded) and consumers are the three type counterparts. The analyse of each of these counterparts can be addressed either using the expertise of a credit risk analyst or adopting a mathematical approach. The objective of credit scoring is to assess the probability that a borrower will show some undesirable behavior in the future \cite{hand}. The nature of available data to estimate the probability of default rely on the counterpart. For publicly traded companies  the literature is focused on two different approaches: the first one starts with the Z-Score \cite{altman}, a models that predict insolvency using historical accounting data. The second approach rely on securities market information (\cite{merton}, \cite{black_cox_str}.

In order to assess the credit risk worthiness of a company, consumer financial and insurance  institutions use financial indicators (i.e., financial ratios computed using accounts information) for business loans, and both personal and financial information for consumer lending. To highlight the relevance of developing a credit score model, \cite{NBERw23740} shows that during the 2007-2009 housing crisis there was a marked rise in mortgage delinquencies and foreclosures among high credit score borrowers, suggesting that credit scoring models at the time did not accurately reflect the probability of default for these borrowers. After the 2008 crisis, the financial institutions became more risk-averse, which provoked a substantial increase on the barriers in the process of acquiring credit \cite{cowling}. 

As shown in  \cite{filosophyreaing} it is important, when developing a PD model or an internal rating, to decide whether to grade borrowers using their current situation or (point-in-time, PIT) or their expected condition over a cycle (through-the-cycle, TTC). Classical Credit Rating Companies use Credit Scorecards to evaluate the risk of a counterpart. This algorithm takes as input financial information and outputs a qualitative estimation of probability of default for a company. 

In the last decades, a growing number of approaches has been developed to model the credit quality of a company by exploring statistical techniques. There are three main generations of statistical techniques \cite{coporate_default_pred}: Discriminant Analysis (\cite{altman}, \cite{beaver}), Binary Response Models (\cite{ohlson}), and Hazard Models (\cite{binary_chava}). 

Machine learning algorithms have shown an increase in the prediction power for Credit Risk Modeling \cite{LESSMANN2015124}.Although they improve the existing credit scoring models, the AI-powered systems are regarded with suspicion because they do not provide reliable explanations for the score they provide. In this context, eXplainable Artificial Intelligence (XAI) \cite{arrieta2019explainable} has rapidly gained interest in the financial field.

In this paper we use historic financial data for predicting the default of a company in one year horizon. We focus on companies based in Europe (mainly France). The information used for modeling the default is shown in \ref{tab:def financial variables}. As we can see in \ref{tab:year distribution}, the dataset is highly imbalanced. We apply several machine learning (ML) techniques as well as resampling techniques to address this problem. Finally, we combine the best model with the SHAP technique \cite{lundberg2017unified}, a XAI method widely used for model interpretation based on feature attribution.

To summarize, these are the contributions of our paper:
\begin{enumerate}
    \item We analyze a large dataset (around 100.000 companies) mainly based in France. Most of the literature on credit scoring algorithms predicts consumers default, while in this work we focus on a large variety of companies (from small companies to big corporations).
    \item We map our machine learning model probabilities to risk score labels in order to compare it with an established companies credit risk scoring system. 
    \item We perform an interpretability study, using the well-known Shapley value analysis (SHAP), in order to understand why the algorithm made a certain decision.
    \item We analyze the most important features of the developed model and compare these features with the expertise of several risk analysts.
\end{enumerate}

\section{Related Work: EXplainable AI for Credit Scoring}
In the last decade the intersection between machine learning and the credit risk community has improved the performance of the credit risk models. In this section we present the main works on credit scoring focusing on machine learning models and non traded companies (i.e., companies which have their shares listed on any stock exchange). Also we show previous studies on eXplainable Artificial Intelligence in finance.


\subsection{Machine learning for credit scoring}
The last decade the relevance of Machine Learning (ML) has grown exponentially across all industries. However the first intersection between finance, in particular credit scoring, and ML industries was in the  80s. Some ML algorithms used in credit scoring are: decision trees \cite{Makowski}, k-nearest neighbors \cite{knn_credit_scoring}, or kernel-based algorithms such as Support Vector Machine (SVM) \cite{svm_baes}. Recently, more sophisticated ML-based models have been applied to credit scoring. In \cite{LESSMANN2015124} they compare a list of 41 different ML models for consumer credit scoring. The results shows the Random Forest Algorithm, a random version of bagged decision trees \cite{breiman2001random}  outperforms the classical and widely used Logistic Regression. 

The scarcity of data for assessing the credit risk of non publicly traded companies has provoked research to be more focused on consumer lending. Nonetheless, there are some works that focus on this subject. In \citet{Bussmann}, they analyze a dataset of companies based in Southern Europe for the year 2015. They use Extreme Gradient Boosting \cite{Chen_2016} for predicting whether a company will default the next year. \citet{provenzano2020machine} build a model to predict the default of a company in a one year horizon using a dataset composed of Italian companies over the period 2011-2017. For PD modeling they use a boosting method called LightGBM \cite{lightgbm}. \citet{MOSCATELLI2020113567} they consider that a company is defaulted for the given year if the ratio of non-performing credits to total credit drawn is greater than 5\%. Their best results has been obtained using Random Forest. 
Companies default as well as consumers default are rare events and thus, when treating with these datasets is important to address potential data collection and reporting bias. In \ref{tab:corporate credit scoring datasets} we summarize the different datasets used in the literature. Several techniques are applied to tackle the imbalance problem . One of the common solutions is generating synthetic data of the minority class (SMOTE) \cite{Chawla_2002}. In \citet{islam2019investigating} they show that applying SMOTE \cite{Chawla_2002} in training stage improves the performance of a large list of ML models for bankruptcy modeling.
\begin{table}[htbp!]
    \centering
    \footnotesize
    \caption{Datasets used for companies credit scoring modeling using machine learning-based models.}
    \label{tab:corporate credit scoring datasets}
    \begin{tabularx}{\linewidth}{l c c X }
    
    \hline
   \textbf{ Dataset Reference} & \textbf{Dataset Size} & \textbf{Features} & \textbf{Imbalance Ratio}\\ \hline
    \citeauthor{Bussmann} \citeyear{Bussmann} \cite{Bussmann} & 15,045 & Not specified & 8.17\\ 
    \citeauthor{risks6020038} \citeyear{risks6020038} \cite{risks6020038} & 117,019 & 181 & 65.67 \\ 
    \citeauthor{provenzano2020machine} \citeyear{provenzano2020machine} \cite{provenzano2020machine} & 919,636 & 179 & 65\\
    \citeauthor{MOSCATELLI2020113567} \citeyear{MOSCATELLI2020113567} \cite{MOSCATELLI2020113567} & $\sim$ 250,000 & 26 & 65\\ 
    Dataset used in this work & 138,419 & 15 (Table \ref{tab:def financial variables}) & 114.75 \\
    \hline
    \end{tabularx}
\end{table}
The imbalance ratio in Table \ref{tab:corporate credit scoring datasets} is the proportion of the number of non defaulted companies to the number of defaulted companies.

\subsection{EXplainable Artificial Intelligence (XAI) in Finance}
The pursuit of highly performant machine learning algorithms has derived in complex systems that are harder to interpret and therefore to trust \cite{Dosilovic2018ExplainableAI}. The challenge for today's ML based credit scoring models and more generally the implementation of AI-powered systems in the financial industry is to meet strong regulations(e.g.,:General Data Protection Regulations (GDPR)). To ensure the correct, ethical and responsible development of AI in finance the implemented systems need to be explainable and interpretable. Recent works \cite{acpr} discuss the requirements an AI-based system needs to meet in order to guarantee the fair functioning of the system. Several techniques have been developed in order to clarify opaque models interpretability problem \cite{arrieta2019explainable,guidotti2018survey}. One of these techniques is SHAP (SHapley Additive exPlanation). SHAP \cite{shap} is a framework used for interpreting predictions based on game theory. It falls into the Post-hoc explainability methods taxonomy of XAI in \cite{arrieta2019explainable}. These family of methods target to explain the output of models that are not readily interpretable by design \cite{arrieta2019explainable}.
Recent works (\cite{xai_ml_credit}, \cite{Demajo_2020}) emphasize the importance of understanding the decision-making process for ML based credit risk models and how addressing this problem could benefit the implementations of more machine learning models in the credit risk industry.

\section{Methodology: Black box models and XAI}
In this section we present the methods we use in our study. First we start by presenting the State-of-the-Art ML models, and then we describe how we prepared our data for the modeling step. We show a procedure to  perform data augmentation and generate synthetic data for the minority class, based on oversampling, in order to improve the performance metrics. We present the different evaluation metrics used to compare the different  ML models. We briefly describe the explainability framework used in this work, SHAP values. In the last stage we detail how we conduct the survey between several credit risk analysts. The survey will be used to compare the results of the explainability framework with the human expert explanations.


\subsection{Companies Credit Risk Scoring Data Preparation} 
\label{sec:data preparation} 
\textit{Data Cleaning:} The data used in this study is provided by Tinubu Square\footnote{\url{https://www.tinubu.com/}}  a company that  provides companies credit risk opinions as a service. Tinubu's database is composed by financial and non financial information about a large set of companies. The initial dataset is composed of 6,051,844 data points and 17 variables. To develop the PD model, we will use the financial variables, therefore, from the original dataset we will keep those evaluations that were made using financial variables. For this study we are interested in those companies for which we have all financial information available. At this stage, we have a total of 1,415,610  assessments (i.e., data points); the number of unique companies assessed is 418,516.

\textit{Data Labeling:} The next step is to create the target variable of our problem. We are considering the problem of modeling the default of a company knowing their previous financial information. Financial information  of a company non publicly traded is published yearly. Since in this study we are interested in short term PD modeling, we fix the time horizon to one year (e.g., given the financial information of a company the year 2012, we want to estimate the probability of default for the year 2013). We need to keep the companies with financial data available for two  consecutive years, knowing that the first year the company has to be a non defaulted company check if the \textit{Out of business} variable in Table \ref{tab:def financial variables} is equal to \textit{No}), and then, if in the second year the company's \textit{Out of business} = \textit{Yes}, then we set the target variable to 1; otherwise, we set the target variable to 0. 

\begin{figure}[htbp!]
    \centering
        \caption{For a given year $t$, the default rate represents the percentage  of defaulted companies in year $t+1$ over all companies rated the year $t$ and $t+1$.}
    \label{fig:default rate}
    \includegraphics[width=\linewidth]{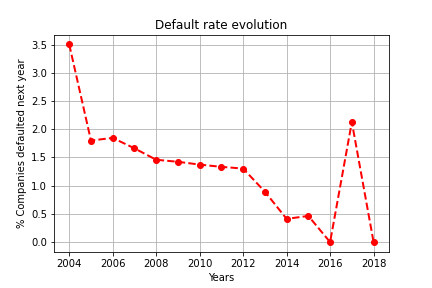}
\end{figure}

In Fig. \ref{fig:default rate}  we observe that the years with around 100K companies, the default rate is in the range [1.5\%-2\%]. During the period between 2015-2018,  the amount of assessed companies is significantly lower than the other years, and consequently the \% of defaulted companies varies heavily.

All original financial variables (see Table \ref{tab:def financial variables}) are important for predicting whether a company will incur into a default. Since the number of missing values is significant (see Table \ref{tab:missing values original variables}), We keep those companies we have all financial variables needed to compute the ratios in Table \ref{Tab:Ratios used to train the models}.

\textit{Data Transformation:}
 First we start by encoding the categorical features. The only categorical variable present in Table \ref{tab:def financial variables} is the country code. We use the one hot encoding technique to create a new column for each country. This new column will take the value 1 if the current company is based in that country and 0 otherwise. The next step is to transform  the original features in Table \ref{tab:def financial variables} by computing  the  ratios  in Table \ref{Tab:Ratios used to train the models}. 

We split the dataset into two main sets: a first set in which we have the data between 2004-2012. We use this data to train and test (70\% for training and 30\% for testing). The data between 2013-2018 will be used  to validate our model.

\textit{Data Normalization:} The transformed data in Table \ref{Tab:Ratios used to train the models} contains noise and had a different scale. Therefore we scale the data using the standard scaler. This helps to reduce the noise by transforming the data distribution into a new one with mean 0 and standard deviation 1.


\begin{figure*}[htbp!]
    \centering
    \caption{Data preparation flowchart including the years over which train and test data are split. }
    \includegraphics[width=0.7\linewidth]{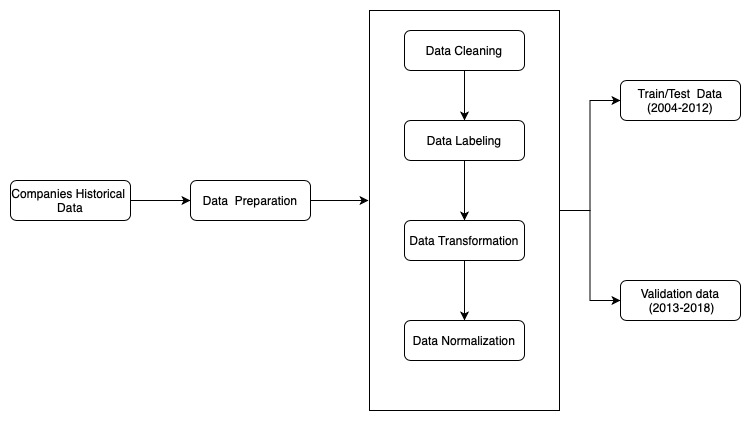}
    \label{fig:flowchart data preparation}
\end{figure*}

\subsection{PD Modeling using Machine Learning Models}
In this part we present the classification models used to predict the probability of default (PD). The output of all models is a probability of default (PD). We reduce our problem to a binary classification model where if this probability is greater or equal to 0.5 then the model is predicting that the company will be in default the next year.  

\textit{Logistic Regression:} a linear model that makes a prediction by computing a weighted sum of the input features. The output of this model is a probability for a binary classification (whether the company will fall into a default or not). This probability can be mapped setting a threshold. The threshold used for predicting whether a company will be in default is 0.5. If the probability is greater than 0.5, then the model predicts the company as defaulted, i.e., bankrupt. Logistic regression has been widely used in the credit risk prediction domain due to its simplicity and interpretability. 

\textit{AdaBoost (AB):} A tree-based ensemble method whose algorithm trains a decision tree and then tries to fix the errors by training sequentially decision trees over the predecessor tree errors.  For the hyperparameter optimization step, we use Grid Search 5 fold cross validation. We obtain the best results with a \textit{Learning rate = 0.8} and \textit{n estimators = 100}.

\textit{Random Forest (RF):} widely used in the machine learning community, the Random Forest algorithm generate trees from a random sample of the training data. For each generated tree, the algorithm randomly selects an attribute for splitting the tree. This randomness reduces the overfitting of the model by decreasing the correlation between trees. The output of a random Forest is a the average of all trees predictions. Between all the parameters tested using Grid Search 5 fold cross validation for the training set, best results (\ref{tab:performance table}) were obtained with 
\textit{n estimators = 1500}.

\textit{Gradient Boosting (XGB):} the principle of operation of the Gradient Boosting lies in the sequential  combination of weak learners to create a robust model. The difference between AdaBoost and Gradient Boost is that the drawback of having weak learners is detected using gradient descent. A variant of the Gradient Boosting, the extreme Gradient boosting has been used in this work. This variant uses a  more regularized model formalization to control overfitting. In Table \ref{tab:performance table} results associated to the XGB model were obtained using the hyperparameters: \textit{learning rate = 0.1}, \textit{n estimators = 100}, \textit{max depth = 10}, \textit{subsample = 1}, \textit{colsample bytree}\textit{gamma = 0.7}. 

All methods employed used the implementations provided by Scikit Learn \cite{scikit-learn}. XGBoost uses the one concretely from \cite{Chen_2016}.

\subsection{Data Oversampling using SMOTE}
The SMOTE oversampling technique \cite{Chawla_2002} consists on oversampling the minority class by taking each minority class sample and introducing synthetic examples along the line segments joining any/all of the k minority class nearest neighbors. Depending upon the amount of over-sampling required, neighbors from the k nearest neighbors are randomly chosen. We resample the training set using SMOTE. We set the parameter k=10 and the ratio between the minority class and majority class in the resampled set to 0.5\footnote{After trying different combinations for both hyperparameters, these values obtain the best results for the tested ML models.}.

\subsection{Performance Metrics}
Measuring the performance of classification problems can be achieved using different metrics. First we begin by comparing the different machine learning models using the following metrics:
\begin{itemize}
    \item \textit{Accuracy}: defined as the ratio of correct predictions 
    \item \textit{Precision}: the proportion of correctly predicted classes over the total of data points. 
    \item \textit{Recall}: the proportion of defaulted companies that are correctly predicted by the model.
    \item \textit{F1-Score}: the weighted average of the precision and recall.
    \item \textit{AUC}: measures model’s ability to discriminate between cases. It is the area under the Receiver Operating Characteristic.
\end{itemize}

\subsection{SHAP for model explanations}
To understand the outputs of the ML model we employ SHAP \cite{shap}, a framework for interpreting model predictions. SHAP uses a game-theoretic approach that explains the contribution of each feature to the final output of a given model. This method will ascertain which financial features are the most relevant in order to predict the default of a company. 

\subsection{Human Expert: Introducing Credit Risk Analysts Expertise}
The main point of this article is to compare the explanations given by the ML model (i.e., in this case the XGBoost) after applying the SHAP \cite{shap} framework with the credit risk analyst opinion. The way this comparison has been conducted is as follows: we asked several Tinubu's risk analysts (to be more precise 4 different risk analysts) to weight what variables are the most important in order to rate a company. We asked them to distribute 100 points between all the variables that has been used for training our ML model. Then we compute the sum of all weights given by the different risk analysts. This sum represents the level of importance for each feature. This value will be used to rank the features by importance degree and it allow us to compare the human expert ranking with the feature importance for our model given by the SHAP value.

\section{Results}
In this section we present the results of the models we described in the previous section for different settings. Then we compare  the results of the best model with Tinubu's Rating System. Finally, we discuss the differences between the decision making process of our ML model and Tinubu's mathematical Credit Rating System.

\subsection{Performance of ML models for PD modeling}

\begin{table*}[h!]
\setlength{\tabcolsep}{8pt}
\renewcommand{\arraystretch}{1}
    \caption{Model's performance using the features in Table \ref{Tab:Ratios used to train the models}. This features are the input of Tinubu's Rating System. WRS corresponds to  training the model  without resampling. RS stands for training the model resampling training data using SMOTE (k=10, R=0.5).  RS+VS represents those models that have been trained with the resampled training set and evaluated over validation dataset.}
    \label{tab:performance table}
\footnotesize

        \begin{tabularx}{\linewidth}{l  X X X X X X}
        \toprule
        Model & \multicolumn{5}{c}{Performance Metrics} \\ \cline{2-6}
        & Accuracy & Precision & Recall & F1-Score & AUC \\ \midrule
        LR (WRS)& \textbf{98.98} & 0 & 0 & 0 & 0.6736\\
        LR (RS) & 97.05 & \textbf{3.58} & 7.44 & 0.0484 & 0.6876\\
        LR (RS+VS) & 97.62 & 0.80 & 4.70 & 0.0137 & 0.7292\\
        AdaBoost (WRS)& \textbf{98.98} & 0 & 0 & 0 & 0.7263\\ 
        AdaBoost (RS) & 89.96 & 3.18 & 28.18 & 0.0505 & 0.7058\\ 
        AdaBoost (RS+VS)& 98.48 & 1.37 & 4.70 & 0.0213 & 0.7324\\ 
        Random Forest (WRS)& \textbf{98.98} & 0 & 0 & 0 & 0.6551 \\
        Random Forest (RS)& 96.71 & 2.82 & 12.85 & 0.0462 & 0.7086 \\
        Random Forest (RS+VS)& 99.64 & 0 & 0 & 0 & 0.6908 \\
        XGBoost (WRS)& \textbf{98.98} & 0 & 0 & 0 & 0.6728 \\
        XGBoost (RS) & 90.35 & 2.92 & \textbf{30.09} & 0.0227 & 0.7027\\
        XGBoost (RS+VS) & 95.39 & 1.22 & 15.29 & \textbf{0.0536} & \textbf{0.7466}\\
        \bottomrule
    \end{tabularx}

\end{table*}

Analyzing Table \ref{tab:performance table}, we observe that the ML models do not recognize the companies that  will default the next year if we train them with the original data. However, there is a improvement for models trained with generated synthetic data of the minority class (i.e., the. defaulted companies). What we notice is that for the validation set, which is composed of companies between 2013-2018 , the XGBoost performs better than all other tested models. The results in Table \ref{tab:performance table} show that the XGBoost model is able to detect defaulted companies with  higher precision (i.e., recall) than the rest of ML models,  
while having similar AUC and Precision.

\subsection{Mapping XGBoost probabilities to Tinubu's grades}
Tinubu's Scorecard Algorithm is the internal proprietary rating system used at Tinubu Square to evaluate the credit risk of a debtor (a company). This rating is a descriptive way to present the probability of default of the assessed entity. The Tinubu's rating scale uses  letters to establish the level of credit worthiness of a company. On one hand, \textit{A} is given for companies that the model estimates the probability of default is close to zero. On the other hand, companies considered very likely to incur into a default are rated with letter \textit{F}. The \textit{X} score is given to companies whose data needed to compute the PD is not available.

\begin{table}[htbp!]
\setlength{\tabcolsep}{8pt}
\renewcommand{\arraystretch}{1}
\footnotesize
\caption{Rating scale for tinubu's algorithm.}
\label{tab:rate scale TA}
    \begin{tabularx}{\linewidth}{c l}
    \hline
         Rating Scale & Probability of default level\\
      \hline
          A & Little/no default risk\\
      
      B & Low default risk \\
      
      C & Average default risk \\
      
      D & Above average default risk \\
      
      E & Increasing and high default risk \\
      
      F & Extremely high default risk or in default \\
      
      X & Excluded companies from the analysis due to lack of data\\

      \hline
    
\end{tabularx}

\end{table}

Our goal is to compare the two models: Tinubu's Scorecard Algorithm and the best machine learning model (i.e., the XGBoost model). Nevertheless the outputs of each both models (Tinubu Scorecard Algorithm  and ML model) are naturally different. As defined above (see Table \ref{tab:rate scale TA}) the Tinubu Scorecard Algorithm  model outputs a category,  while the ML model outputs a continuous variable (probability of default). To compare both models we need to create a mapping to assign to each score letter a probability of default. As we consider the gold standard the financial experts model of Tinubu our ground truth, this PD will need to be matched by the ML model output. 
The proposed  mapping  will allow us compare the Tinubu Scorecard Algorithm with the ML model (i.e., the XGBoost model): 

Given a Tinubu's score class we compute the average of the probabilities of the companies being in default next year (i.e., $Y_t = 1$).
\begin{equation}
    \mu(R)=\frac{1}{N}\sum_{i=0}^{n-1} \mathbb{P}_i(Y_{t+1} =1\mid Rate=R) \; \; \textrm{where} \; \; R\in[A-F] 
    \label{eqn: ecaucion mean probabilities}
\end{equation}
where $R$ is the rate yielded by Tinubu's Scorecard Algorithm. $\mu(R)$ represents the mean of the probabilities of default given by the XGBoost model when the yielded rate by the Tinubu's Scorecard is $R$.
 
\begin{equation}
    \label{argmin}
        \hat{R} = \textit{argmin}_{R} \Big \{ \mathbb{P}(Y_{t+1} \mid X_t) -\mu(R) \Big \}
\end{equation}
Then for each individual company we estimate the rating by searching for the Rate that minimizes  the difference between  the PD of the company given by the ML model and the average probability estimated  by the ML model of all different Tinubu's Rating classes (see equation \ref{argmin}). The estimated rating $\hat{R}$ represents the rating yielded by the XGBoost after the mapping process.

\begin{table}[htbp!]
\setlength{\tabcolsep}{8pt}
\renewcommand{\arraystretch}{1}
\footnotesize
\caption{Scorecard mapping: we assign a Tinubu-defined credit risk score to each probability yield by XGBoost (for which we compute the interval bounds thanks to Eq. \ref{eqn: ecaucion mean probabilities} and \ref{argmin}).}
\label{tab:year distribution}
    \begin{tabularx}{\linewidth}{X X X}
    \hline
        Tinubu Score & Mapped to interval $\mathbb{P}(Y_{t+1}=1)$ (Interval lower bound) & $ \mathbb{P}(Y_{t+1}=1)$ (Interval Upper Bound):\\
      \hline
      A & 0 & 0.0828 \\
      
      B & 0.0828 & 0.1411 \\
      
      C & 0.1411  & 0.2029 \\
      
      D & 0.2029 & 0.2486\\
      
      E & 0.2486 & 0.285\\
      
      F & 0.285 & 1\\

      \hline
    
    \end{tabularx}
\end{table}
\begin{figure}[hbtp!]
    \centering
    \caption{
    Confusion matrix for mapping Tinubu's 
    Scorecard algorithm risk labels to the ML model probabilities. The objective is to assign a probability prototype representing each letter rank used by Tinubu's PD classification problem, and evaluate our ML model with the test set. The columns refers to the rating given by the XGBoost model while the the rows represent Tinubu's rating. We observe that our XGBoost model is conservative if we compare it with the Tinubu Scorecard Algorithm (e.g., the number of companies with a  a F grade for XGBoost model is higher than the companies rated F by the Tinubu's Scorecard Algorithm. Large part of the companies rated F by the XGBoost falls into the range [\textit{C}-\textit{E}])\ref{tab:rate scale TA}.}
    \includegraphics[width=\linewidth]{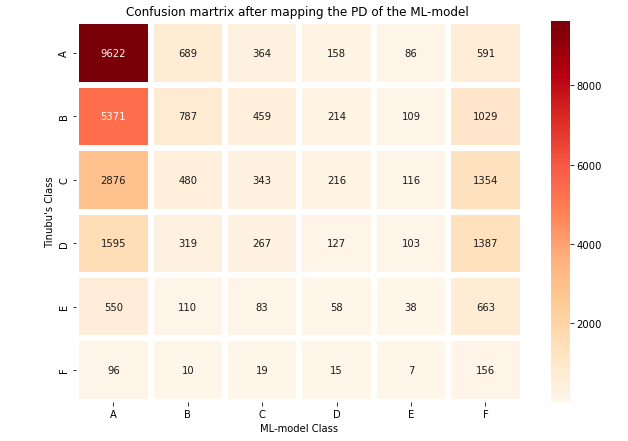}
    \label{fig:mapping pd}
\end{figure}

In Figure \ref{fig:mapping pd} we observe that the ML model is able to assign correctly low risk labels to companies (A and B labels). However, the ML model overestimates the risk of a significant number of companies. This is shown in Table \ref{fig:mapping pd} in column F. In the context of Credit Scoring is important to remark the impact of underestimating the risk. The critical case occurs when the model estimates that a company has a low or near zero risk of being in default the year after the assessment, and the reality is that that company is much riskier than considered by the model. This is why we consider that the XGBoost model  works relatively well, in comparison with a well established Rating Algorithm (Tinubu's Scorecard Algorithm) that has been rating companies since early 2000s and is still being used nowadays.

\subsection{Explaining our PD model: SHAP value analysis}

\begin{figure*}[hbtp!]
    \centering
    \caption{Contribution of each explanatory feature to the final prediction based on Shapley analysis of contribution decomposition for the default prediction. Axis x represents whether the value of the feature contributes positively (negative values in x axis reduce the probability of default and viceversa).While some features contribution are easy to interpret, since high or low values are homogenized and concentrated in one range of the horizontal plot in a single colour, others are harder. When this is not the case, it means we can not conclude how high (or low) values of these variables affects the probability of default when talking about a high (or low) value of such feature in a general manner. Features are sorted according to the their relevance (i.e., SHAP average absolute value).}
    \includegraphics[width=0.8\textwidth]{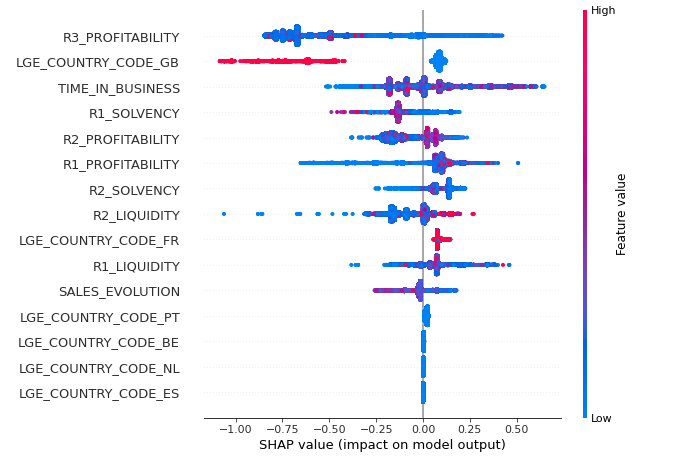}
    \label{fig:shap analysis}
\end{figure*}
Previous sections provided evaluation metrics for ML models attempting to reach performance levels of years of experience of financial risk analysts using mathematical models at the core of their business. 
In this part we analyze how the ML model has arrived to the results shown in Tables \ref{tab:performance table} and \ref{fig:mapping pd}.

Features in Fig. \ref{fig:shap analysis} are ranked from the most relevant (R3 profitability ratio) to the less important ratio (country code being from Spain). This analysis highlights the fact that, for the ML model (i.e., the XGBoost model), companies based in Great Britain have a lower probability of default than companies based in other analyzed countries (France, Belgium, Spain, Netherlands, Portugal). This fact matches with the way Tinubu's Algorithm works. For companies with financial ratios relatively similar, the Tinubu's Scorecard Algorithm  will give a better rating to those companies based in Great Britain. The contribution of a given feature whose data points lie next to other data points of opposite colour in the same axis show that both high and low values of that feature have influenced similarly the model outcome, and thus their contribution may be contextually better determined by considering generic values of other out-weighting features that have stronger contribution for that data point.

\subsection{Feature Contribution Analysis: Assessing explanations from risk analyst experts vs ML models}
In order to assess the explainability of our ML model with the respective mathematical model of Tinubu Square, we asked 4 different analysts from Tinubu Square to give a weight for all features in Table \ref{Tab:Ratios used to train the models}. Each analyst has 100 points to distribute them between the features. All participants of this survey filled the Table independently. As all analysts work at Tinubu Square, we expect a relatively uniform criteria.  Comparing the weights  given by the analysts in Table \ref{Tab:contribution according to risk analysts}, and the importance of each feature  for the ML model (see SHAP analysis in Fig. \ref{fig:shap analysis}), we conclude that the decision making process clearly diverges among machine and human models. Generally, the assessment of the credit worthiness of a given company is made by a human expert if the amount of credit demanded by the assessed company is relevant. Therefore, the dataset of companies treated by the risk analyst may be biased towards companies that can afford larger interest rates or higher insurance premiums.

For all analysts the \textit{$R_3$ Profitability} feature is irrelevant, while for our ML model is the most important one. Analyzing the most important features for both the human-expert and the ML model we discover that the only feature considered relevant when assessing a company credit worthiness for both human experts and the ML model is the \textit{$R_1$ Solvency}, which measures the ability of a company to meet short-term debts. From a credit risk analyst stand point, the most important features are those related to short-term activities (i.e., \textit{$R_1$ Liquidity}, \textit{$R_2$ Liquidity} and \textit{$R_2$ Solvency}). However for our ML model, the 
features that contribute the most to the model output are the features that relate to activities extended in time, i.e., long term capabilities of the firm, for instance, \textit{$R_3$ Profitability}, \textit{Time in Business} and \textit{location}.

\begin{table*}[hbtp!]

\setlength{\tabcolsep}{8pt}
\renewcommand{\arraystretch}{1}
\footnotesize
\centering
 \caption{Weight of each feature given by 4 different risk analysts at Tinubu Square. The risk analyst distribute 100 points between all the features. Features are ordered by decreasing importance for the risk analysts (i.e., total points the feature has received from all experts). The sum of weights given by the analysts is the same for $R_2$ Solvency and $R_2$ Profitability. We ranked  $R_2$ Solvency  above $R_2$ Profitability because the weights are similar.}
 \label{Tab:contribution according to risk analysts}

\begin{tabularx}{\linewidth}{l X X X X X }
     \hline
     \textbf{Features} & \textbf{Risk Analyst Expert 1} & \textbf{Risk Analyst Expert 2}& \textbf{Risk Analyst Expert 3}&\textbf{Risk Analyst Expert 4} & \textbf{Total Weights}\\ 
     \hline
     $R_2$ Liquidity & 20 & 30& 30& 20  & 90\\
    
        $R_1$ Solvency & 25 & 10 & 25 &  20 & 80\\
        
        $R_2$ Solvency & 5 & 10 & 25 & 15 & 55\\
        
        $R_2$ Profitability & 10 & 30& 5& 10 & 55\\

        $R_1$ Liquidity &  15 & 10 & 15 & 5& 45 \\
        
        Sales evolution &   5 & 3& 1& 10 & 19  \\
        
        Country code  &  10 & 2& 2& 5 & 19\\
        
        Time in business &    5 & 5 & 2 & 5 & 17 \\
        
        $R_1$ Profitability &  5 & 0& 5& 5 & 15 \\
        
        $R_3$ Profitability  &    0 & 0 & 0&  5 & 5 \\
        
\hline
\end{tabularx}%
\end{table*}
In Table \ref{Tab:contribution according to risk analysts} we rank the features relevance depending on the risk analysts. The latter act as a proxy of the gold standard credit risk opinions. Their expertise is captured by the Tinubu internal Scorecard Algorithm.

\section{Discussion}
The first issue concerns the usage of a highly imbalanced  dataset. It is worth noting that the task of predicting the probability of default entails having to deal with bias inherent to the nature of the data, since the percentage of default companies is very low with respect to successful companies that do not fall on bankruptcy (around 1 \% of defaulted companies the next year after the assessment). Therefore the ML model may have problems  differentiating between both defaulted and non defaulted companies. Since defaults are very rare events, resampling data by generating new synthetic data  of the minority class (SMOTE) may not improve models performance because each event of default has 
a particular context and the concept of default may differ from one country to another. It is worth highlighting the comparison between the Tinubu's Scorecard algorithm and the ML model (i.e., the XGBoost model). This shows that the ML model does 
underestimate the number of highly ranked firms (i.e., ranked with top score \textit{A, B} and \textit{C} by Tinubu's Scorecard Algorithm). On the other hand, we observe that the ML (XGBoost) model rates a considerable number of companies with the \textit{F} score, while the Tinubu Scorecard Algorithm maps the same companies mainly to scores in the range [\textit{B}-\textit{D}]. This behaviour can be interpreted as the ML model being significantly more conservative than the Tinubu's Scorecard Algorithm. The property of being risk averse is therefore desirable in our context, and thus, beneficial due to safety reasons. 

\section{Conclusion}
In this work a number of state-of-the-art machine learning models have been proposed to model the probability of default of a company the year after it has been assessed. For the Tinubu dataset used, we remark that due to the relatively high imbalance ratio (see Table \ref{tab:corporate credit scoring datasets}) in comparison with other companies datasets used in the credit risk scoring literature, the ML problems do not 
distinguish well? which companies will suffer to meet their credits. 
The best result among all implemented models reports an AUC of 0.7466 (Corresponding to 95.39 \% accuracy), and was achieved with the XGBoost model.
However, using techniques for data resampling before the training stage improves the model performance recall metric up to 20\%.  In this context, future works should focus on treating the imbalance problem by trying different data augmentation techniques and models.

In order to evaluate our ML model's PD, casted as a regression problem, with risk analyst experts categorical scores predicting a PD in form of a score, we mapped the probabilities given by 
the best ML model (i.e., XGBoost, as it showed best results) 
to Tinubu's Scorecard model score labels. This mapping shows that the ML model is able to 
tell apart companies with low risk of default (i.e., companies rated with an \textit{A} and \textit{B} with the Tinubu's Scorecard Algorithm) 
from the rest of companies. 

One of the biggest challenges for the introduction of machine learning based models in the credit scoring field, in particular for companies credit scoring, is the lack of credibility, trust, and explainability. We addressed this problem by using the explainable framework of SHAP analysis. 
Assessing the results of the SHAP analysis, we conclude that the difference between companies based on the UK and the rest of the companies (based on Southern Europe) may require a deeper analysis, since the hypothesis of different  country regulations could affect the companies default analysis.

The main contributions of this work consist of a study analyzing explanations given by a SHAP analysis of feature contributions with respect to explanations backed up by the expertise of a pool of credit analysts. The comparison showed that the criteria between human experts and the ML model are quite different. In particular:

\begin{enumerate}
\item The ML model excels at being able to capture longer term abilities of a company, with respect to the features considered by the Tinubu risk analyst experts. The latter focus more on shorter term variables such as \textit{$R_1$ Liquidity}, or \textit{$R_2$ Liquidity}, while the ML model attributes higher relevance to life-long attributes of the firm.
\item We found interesting the fact that, without explicitly implementing any constraint, XGBoost arrives to the conclusion that companies that exercise their activities in Great Britain are more likely to avoid financial problems, 
while this behaviour is explicitly encoded and accounted 
by Tinubu's Scorecard Algorithm. 
\item Since credit risk scoring algorithms, specially those focused on companies rather than individuals, are in the very early stages, we acknowledge the abilities of our ML model to remain conservative when estimating risk. This is a desirable property of 
such complex models, since it is preferable to avoid critically large economical losses. 

\end{enumerate}
The most critical result is that for our ML model, the most relevant feature is $R_3$ Profitability, a ratio that measures how effectively a company is using its assets to generate earnings, while all consulted experts (i.e., Tinubu's risk analysts) consider this feature irrelevant. This fact highlights the importance of focusing not just on being able to build explainable ML models but creating a consensus between the human expert and the machine, as well as among experts (which is not always the case).

Future works should focus on studying how inductive biases can infuse expert knowledge into the ML model, for instance, by introducing credit expert opinions and preferences (at the data annotation and model design stages) in order to improve the model performance. Other potential avenue of research is designing a sound basis for causal explainability that can be verified and certified by human experts. Moreover, it may be interesting to further analyze the geographical locality context, i.e., contextually and historically, by assessing companies both by risk analysts and the ML model.
Finally, we hope future work designs explainability metrics to programmatically asses the quality of an explanation given by a black box model that can continually learn, evolve and degrade over time, to assess its fidelity, i.e., alignment, with human expert opinions. This way we will be able to deploy AI systems that both human and experts can mutually improve, support, and trust.


%

\section*{Acknowledgment}

The authors would like to thank the Tinubu team of experts for providing help in rating the financial scores.

\ifCLASSOPTIONcaptionsoff
  \newpage
\fi



%



\bibliographystyle{IEEEtranN} 
\bibliography{biblio}

%





\onecolumn
\appendix


\section{Original financial variables provided by Tinubu Square.}
\begin{table*}[htbp!]
\setlength{\tabcolsep}{8pt}
\renewcommand{\arraystretch}{1}
\small
    \caption{Definition of original financial variables. This variables are used to compute the financial ratios in table\ref{Tab:Ratios used to train the models}.}
    \begin{tabular}{p{0.3\linewidth}p{0.6\linewidth}}
\toprule
        \textbf{Feature name} &  \textbf{Feature Description}\\\midrule
        LGE ID &  Identification number. This value is unique for each company\\ 

Statement date  & Corresponds to the date in which the financial data was published \\

Out of business indicator & Binary variable: Yes if the company is currently defaulted\\

Country code & Abbreviation of the country in which the company is based \\

Total Employees & Number of employees  \\

Net worth & Total amount of Equity \\

Total Assets & Refers to the total amount of assets owed by the entity \\

Gross Income & Amount of money earned before taxes\\

Total Liabilities & Combined debts a company owes \\

Current Ratio & A liquid ratio that measures the ability of a company to pay short-term\\

Cash and Liquid Assets & Refers to assets that can be readily convert to cash  \\

Sales & Net sales for the period after returns, allowances, and discounts are deducted\\
Working Capital & Capital of the financial activity period\\

Net Income & Amount left over after all expenses and taxes are deducted \\

Incorporation Year & The year the business incorporated \\

Previous Sales & Financial statement date \\

\bottomrule

    \end{tabular}

    \label{tab:def financial variables}
\end{table*}

\subsection{Distribution of missing values data by year and variable}
\begin{table*}[htbp!]
\setlength{\tabcolsep}{8pt}
\renewcommand{\arraystretch}{1}
\footnotesize
\caption{Percentage of missing values for each financial variable by year.}
    \begin{tabular}{p{0.15\linewidth}lllllllllll}
    \toprule
    Financial Variable & 2005 & 2006 & 2007 &2008 &2009 &2010 &2011 &2012 &2013 &2014\\
\midrule
Total Employees & 6.23  &  6.1 & 10.5&21.22 &25.51 & 27.94& 32.05& 38.37& 37.76& 35.6\\

Net worth & 0.77 &  6.94&7.94 &5.57 &0.97 &0.54 &0.71 & 0.23&0.1 &0.07\\

Total Assets & 0.84  & 6.93&7.61 &5.58 & 1.03&0.66 & 1.13& 0.85& 0.79&0.78\\

Gross Income & 14.74 & 14.35&18.84 &13.31 &8.99 & 12.4& 13.13& 12.03&13.78 &13.75\\

Total Liabilities & 0.87 &  6.97&7.49 &5.27 & 1.18& 0.86 &1.37 & 1.06&1.04&1.01\\

L1 Ratio & 77.59 & 82.23&77.82 &80.47 &40.12 &11.43 &6.47 &4.43 & 4.61&4.04\\

Cash and Liquid Assets & 4.44 &  10.74&11.12 & 8.95&4.48 & 4.11&4.53 &4.47 & 4.28&4.03\\

Sales & 12.5 & 8.7&11.79 &10.02 & 6.72& 8.82& 10.92& 9.1& 10.12&10.47\\

Working Capital & 7.79 & 13.27&30.49 & 52.29& 58.9& 60.1& 63.7& 69.33& 70.87&71.69\\

Net Income & 6.37  & 10.16&13.76 & 10.53& 6.07& 8.95&9.15 & 7.71& 8.5&8.76\\

Incorporation Year & 16.16 & 6.34& 0.83&0.27 & 0.39& 0.51&0.5 &0.48 & 0.36&0.27\\

Previous Sales & 18.53  &  14.99& 18.49&17.36 & 12.3& 15.2&14.4 & 11.74&12.52 &12.41\\
\bottomrule

    \end{tabular}
    
    \label{tab:missing values original variables}
\end{table*}
\newpage
\subsection{Finacial Ratios used for modeling the Probability of Default}

\begin{table*}[htbp!]

\setlength{\tabcolsep}{8pt}
\renewcommand{\arraystretch}{1.2}
\small
 \centering

\vspace{4.5mm}
\caption{Financial ratios used in the financial industry. This ratios are the inputs of Tinubu's Scorecard Rating Algorithm and the ML models used to predict the default of a given company. Tinubu outputs a score A-F, while the ML model outputs a probability of default at one temporal year horizon. Several deterministic mappings exist. The one chosen in our experiments is in Table \ref{tab:mapping PD} }
\begin{tabular}{lcp{0.33\linewidth}}
     \hline
     \textbf{Features} &\textbf{Definition} & \textbf{Description}\\ 
     \hline
     
        $R_1 $ Solvency &$\frac{\textrm{Net Worth}}{\textrm{Total Assets}}$& Measures enterprise ability to meet current debt obligations. High $S_1$ values is indicative of greater solvency\\
        
        $R_2$ Solvency &$\frac{\textrm{Financial Debt}}{\textrm{Gross Income}}$ & Represents the percentage of the gross income that goes to debt payments\\
        
        $R_1$ Liquidity &$\frac{\textrm{Total Current Assets}}{\textrm{Total Current Liabilities}}$& The current ratio measures the ability to pay short term obligations (within one year)\\
        
        $R_2 $ Liquidity & $\frac{\textrm{Cash Liquid Assets}}{\textrm{Sales}}$ & Liquidity indicator that represents the percentage of liquid assets over the revenues of the company\\
        
        $R_1$ Profitability &  $\frac{\textrm{Working Capital}}{\textrm{Sales}}$& Shows the relationship between the funds used to finance company's activities and the revenues a company generates as a result\\
        
        $R_2$ Profitability &  $\textrm{Net Income}$
        & Is an indicator of company's profitability\\
        
        $R_3$ Profitability  &  $\frac{\textrm{Gross Income}}{\textrm{Total Assets}}$& Measures how effectively a company is using its assets to generate earning\\
        
        \textrm{Time in business} &  $\textrm{Assessment year - incorporation year}$ & Years in business\\
        
        $\textrm{Sales evolution}$ & $\textrm{Current sales  - previous year sales}$ & Measures the sales evolution\\
        
        Country code  & $\textrm{Country codification}$ & Country abbreviation in which the company is located  \\
        
\hline
\end{tabular}%


 \label{Tab:Ratios used to train the models}%
\end{table*}

\end{document}